%% file: StableDINO.tex
\documentclass[10pt,twocolumn,letterpaper]{article}

\usepackage{iccv}
\usepackage{times}
\usepackage{epsfig}
\usepackage{graphicx}
\usepackage{amsmath}
\usepackage{amssymb}
\usepackage{nccmath}
\usepackage{caption}
\usepackage{subcaption}
\usepackage{xspace,mfirstuc,tabulary}
\usepackage{tabu}
\usepackage{url}
\usepackage{booktabs}
\usepackage{multirow}
\usepackage{wrapfig}
\usepackage{comment}
\usepackage{enumitem}
\usepackage{accents}

\usepackage{booktabs}
\usepackage{pifont}
\newcommand{\cmark}{\ding{51}}%
\usepackage[table,x11names, dvipsnames]{xcolor}

\newlength\savewidth\newcommand\shline{\noalign{\global\savewidth\arrayrulewidth
  \global\arrayrulewidth 1pt}\hline\noalign{\global\arrayrulewidth\savewidth}}
\newcommand\paperurl[1]{{\footnotesize{\color{blue}{\url{#1}}}}}

\newcommand{\blue}[1]{\textcolor{Cerulean}{#1}}
\newcommand{\green}[1]{\textcolor{OliveGreen}{#1}}

\usepackage[breaklinks=true,bookmarks=false]{hyperref}

\iccvfinalcopy 

\def\iccvPaperID{9211} 

\ificcvfinal\fi

\begin{document}

\title{Detection Transformer with Stable Matching}

\author{\textbf{Shilong Liu}\textsuperscript{\rm 1,2}\thanks{Equal contributions. List order is random.} \thanks{This work was done when Shilong Liu, Hao Zhang, Feng Li, and Hongyang Li were interns at IDEA.},\; \textbf{Tianhe Ren}\textsuperscript{\rm 2}$^{\mathrm{*}}$, \textbf{Jiayu Chen}\textsuperscript{\rm 3}$^{\mathrm{*}}$, \\
\textbf{Zhaoyang Zeng}\textsuperscript{\rm 2}, \textbf{Hao Zhang}\textsuperscript{\rm 2,4}, \textbf{Feng Li}\textsuperscript{\rm 2,4}, \textbf{Hongyang Li}\textsuperscript{\rm 2,5},  \\
\textbf{Jun Huang} \textsuperscript{\rm 3}, \textbf{Hang Su}\textsuperscript{\rm 1}, \textbf{Jun Zhu}\textsuperscript{\rm 1\ddag}, \textbf{Lei Zhang}\textsuperscript{\rm 2}\thanks{Corresponding authors.}. \vspace{0.2cm} \\
    \textsuperscript{\rm 1} Dept. of Comp. Sci. and Tech., BNRist Center, State Key Lab for Intell. Tech. \& Sys., \\
    Institute for AI, Tsinghua-Bosch Joint Center for ML, Tsinghua University \\
    \textsuperscript{\rm 2} International Digital Economy Academy (IDEA) \quad 
    \textsuperscript{\rm 3} Platform of AI (PAI), Alibaba Group \\
    \textsuperscript{\rm 4} The Hong Kong University of Science and Technology \\
    \textsuperscript{\rm 5} South China University of Technology \\
\centerline{  \small
liusl20@mails.tsinghua.edu.cn
\quad \{rentianhe, zengzhaoyang\}@idea.edu.cn   
\quad \{yunji.cjy, huangjun.hj\}@alibaba-inc.com } \vspace{-0.1cm}\\
\centerline{ \small
\{hzhangcx, fliay\}@connect.ust.hk
\quad ftwangyeunglei@mail.scut.edu.cn
\quad \{suhangss, dcszj\}@mail.tsinghua.edu.cn
 \quad leizhang@idea.edu.cn
}
}

\maketitle
\ificcvfinal\thispagestyle{empty}\fi

\begin{abstract}
    This paper is concerned with the matching stability problem across different decoder layers in DEtection TRansformers (DETR). 
    We point out that the unstable matching in DETR is caused by a \emph{multi-optimization path} problem, which is highlighted by the one-to-one matching design in DETR. 
    To address this problem, we show that the most important design is to \emph{use and only use} positional metrics (like IOU) to supervise classification scores of positive examples.
    Under the principle, we propose two simple yet effective modifications by integrating positional metrics to DETR’s classification loss and matching cost, named position-supervised loss and position-modulated cost. 
    We verify our methods on several DETR variants. Our methods show consistent improvements over baselines. By integrating our methods with DINO, we achieve $50.4$ and $51.5$ AP on the COCO detection benchmark using ResNet-50 backbones under $1\times$ (12 epochs) and $2\times$ (24 epochs) training settings, achieving a new record under the same setting. We achieve $\textbf{63.8}$ AP on COCO detection test-dev with a Swin-Large backbone.
    Our code will be made available at \url{https://github.com/IDEA-Research/Stable-DINO}.
\end{abstract}

\section{Introduction}
\label{sec:intro}

\input{images_tex/performance_compare}

Object detection is a fundamental task in vision with wide applications. Great progress has been made in the last decades with the development of deep learning, especially the convolutional neural network (CNN) \cite{faster-rcnn, gao2021fast, ge2021yolox, dai2021dynamic}. 

Detection Transformer (DETR) \cite{carion2020end} proposed a novel Transformer-based object detector, which attracted a lot of interest in the research community. It gets rid of the need for all hand-crafted modules and enables end-to-end training. 
One key design in DETR is the matching strategy, which uses Hungarian matching to one-to-one assign predictions to ground truth labels.
Despite its novel designs, DETR also has certain limitations associated with this innovative approach, including slow convergence and inferior performance.
Many follow-ups tried to improve DETR from many perspectives, like introducing positional prior \cite{meng2021conditional, anchordetr, liu2022dabdetr, gao2021fast}, extra positive examples \cite{li2022dn, chen2019hybrid, groupdetr}, and efficient operators \cite{zhu2020deformable, DuyKienNguyen2023BoxeRBF}. With many optimizations, DINO \cite{zhang2022dino} set a new record on the COCO detection leaderboard, making the Transformer-based method become a main-stream detector for large-scale training.

Although DETR-like detectors\footnote{We focus on DETR-like models with Huagurian matching for label assignments in this paper.} achieve impressive performance, one critical issue that has received insufficient attention to date, which may potentially compromise the model training stability.
This issue pertains to the \textit{unstable matching} problem across different decoder layers. 
DETR-like models stack multiple decoder layers in the Transformer decoder.
The models assign predictions and calculate losses after each decoder layer. However, the labels assigned to these predictions may differ across different layers. This discrepancy may lead to conflict optimization targets under the one-to-one matching strategy of DETR variants, where each ground truth label is matched with only one prediction.

\input{images_tex/multi_optimization_path}

To the best of our knowledge, only one work \cite{li2022dn} has attempted to address the issue of unstable matching problem to date.
DN-DETR \cite{li2022dn} proposed a novel de-noising training approach by introducing extra hard-assigned queries to avoid mismatching. 
Some other work \cite{HDETR, groupdetr} added extra queries for faster convergence but did not focus on the unstable matching problem.
In contrast, we solve this problem by focusing on the matching and loss calculation process\footnote{We have tried some direct but useless solutions for the problem, which will be shown in Sec. \ref{sec:more_stable}.}. 

We present that the key to the unstable matching problem is the \textit{multi-optimization path} problem. 
As shown in Fig. \ref{fig:multi_optimization_path}, there are two imperfect predictions during training.
Prediction A has a higher Intersection over Union (IoU) score but a lower classification score, while prediction B is the opposite. This is the simplest but most common case during training. 
The model will assign one of them to the ground truth, resulting in two optimization preferences: one that encourages A, which means encouraging predictions with high positional metrics to get better classification results, and the other is to encourage B, which means encouraging predictions with high semantic metrics (classification scores here) to get better IOU scores. 
We refer to these preferences as different optimization paths.
Due to the randomness during training, each prediction has a probability of being assigned as a positive example, with the other being viewed as a negative example. 
Given the default loss designs, whether A or B is selected as the positive example, the model will optimize it towards its alignment with the ground truth bounding box, which means the model has \textit{multi-optimization paths}, as shown in the right table in Fig. \ref{fig:multi_optimization_path}.
This issue is less significant in traditional detectors, as multiple queries will be selected as positive examples. However, the one-to-one matching in DETR-like models magnifies the optimization gap between predictions A and B, which makes model training less efficient. 

To solve the problem, we find the most critical design is to \textit{use and only use} positional metrics (e.g., IOU) to supervise the classification scores of positive examples. More formal presentations are available in Sec. \ref{sec:IOU_loss}. 
If we use position information to constrain classification scores, the prediction B will not be encouraged if it is matched since it has a low IoU score. As a result, only one optimization path will be available, mitigating the \textit{multi-optimization paths} issue. 
If extra classification score-related supervision is introduced, the \textit{multi-optimization path} will still impair the model performance, since the prediction B has a better classification score.
With this principle, we propose two simple but effective modifications to the loss and matching cost: position-supervised loss and position-modulated cost. Both of them enable faster convergence and better performance of models.
Our proposed approach also establishes a link between DETR-like models and traditional detectors, as both encourage predictions with high positional scores to have better classification scores. More detailed analyses are available in Sec. \ref{sec:analysis}.

Moreover, we have observed that fusing the backbone and encoder features of models can facilitate the utilization of the pre-trained backbone features, leading to faster convergence, especially in early training iterations, and better performance of models with nearly no extra costs. We propose three fusion ways and empirically select the dense memory fusion for the experiments. See Sec. \ref{sec:memry_fusion} for more details.


We verify our methods on several different DETR variants. Our methods show consistent improvement in all experiments. 
We then build a strong detector named \textbf{Stable-DINO} by combining our methods with DINO. Stable-DINO presents impressive results on the COCO detection benchmark. The comparison between our model and other DETR variants is shown in Fig. \ref{fig:model_comparison}.  Stable-DINO achieves $50.4$ and $51.5$ AP with four feature scales from a ResNet-50 backbone under $1\times$ and $2\times$ training schedulers, with $+1.4$ and $+1.1$ AP gains compared with DINO baselines. With a stronger backbone Swin Transformer Large, Stable-DINO can achieve $57.7$ and $58.6$ AP with $1\times$ and $2\times$ training schedulers. To the best of our knowledge, these are the best results among DETR variants under the same settings.


\section{Stable Matching}

This section presents our solution to the unstable matching problem in DETR-like models.
We first review the loss functions and matching strategies in previous work (Sec. \ref{sec:review_detr}). To solve the unstable matching problem, we demonstrate our modifications on losses and matching costs in Sec. \ref{sec:IOU_loss} and Sec. \ref{sec:IOU_matching}, respectively.

\subsection{Revisit DETR Losses and Matching Costs}
\label{sec:review_detr}

Most DETR variants \cite{carion2020end, meng2021conditional, anchordetr, liu2022dabdetr, li2022dn, zhang2022dino, zhu2020deformable} have a similar loss and matching design. We use the state-of-the-art model DINO as an example. It inherits loss and matching from Deformable DETR \cite{zhu2020deformable} and the design is commonly used in DETR-like detectors \cite{zhu2020deformable, meng2021conditional, liu2022dabdetr, li2022dn, adamixer22cvpr}. Some other DETR-like models \cite{carion2020end} may use a different design but with only minor modifications.

The final losses in DINO are composed of three parts, a classification loss $\mathcal{L}_{cls}$, a box L1 loss $\mathcal{L}_{bbox}$, and a GIOU loss $\mathcal{L}_{GIOU}$ \cite{rezatofighi2019generalized}. The box L1 loss and GIOU loss are used for object localization, which will not be modified in our model. We focus on the classification loss in the paper. 
DINO uses the focal loss \cite{lin2017focal} as the classification loss:

\begin{equation}
    \label{eq:focal}
    \mathcal{L}_{cls} = \sum_{i=1}^{N_{pos}}{|1-p_i|^{\gamma} \mathrm{BCE}(p_i, 1)} + \sum_{i=1}^{N_{neg}} p_i^{\gamma} \mathrm{BCE}(p_i, 0),
\end{equation}

\noindent
where $N_{pos}$ and $N_{neg}$ are the number of positive and negative examples, $\mathrm{BCE}$ means binary cross-entropy loss, the $p_i$ is the predicted probability of the $i$\textsuperscript{th} example, the $\gamma$ is a hyperparameter for focal losses, and the notation $|\cdot|$ is used for absolute value.

A matching process determines the positive and negative examples. Typically, a ground truth will be assigned only one prediction as the positive example. Predictions with no ground truths assigned will be viewed as negative examples. 

To assign predictions with ground truths, we first calculate a cost matrix $\mathcal{C} \in \mathbb{R}^{N_{pred}\times N_{gt}}$ between them. The $N_{pred}$ and $ N_{gt}$ are the number for predictions and ground truths. Then a Hungarian matching algorithm will perform on the cost matrix to assign each ground truth a prediction by minimizing sum costs.

Similar to the loss functions, the final cost includes three items, a classification cost $\mathcal{C}_{cls}$, a box L1 cost $\mathcal{C}_{bbox}$, and a GIOU cost $\mathcal{C}_{GIOU}$ \cite{rezatofighi2019generalized}. We focus only on the classification cost as well.
For the $i$\textsuperscript{th} prediction and the $j$\textsuperscript{th} ground truth, the classification cost is:

\begin{equation}
    \label{eq:cls_cost}
    \mathcal{C}_{cls}(i, j) = {|1-p_i|^{\gamma} \mathrm{BCE}(p_i, 1)} - p_i^{\gamma} \mathrm{BCE}(1-p_i, 1).
\end{equation}

\noindent
The formulation is similar to the focal cost but has a litter modification\footnote{We formulate the implementations of Deformable DETR (\url{https://github.com/fundamentalvision/Deformable-DETR/blob/main/models/matcher.py\#L79-L81}) and DINO (\url{https://github.com/IDEA-Research/detrex/blob/main/detrex/modeling/matcher/matcher.py\#L132-L134}).}. The focal loss only encourages positive examples to predict $1$, while the classification cost adds an additional penalty term to avoid it to $0$.

\subsection{Position-Supervised Loss}
\label{sec:IOU_loss}

To solve the multi-optimization problem, we only\footnote{The ``only'' means that the $f_1(\cdot)$ function in Eq. \ref{eq:new_loss} is related to positional metrics only.} use a positional score to supervise the training probabilities of positive examples. Inspiring by previous work \cite{TOOD, GFL}, we can simply modify the classification loss Eq. \ref{eq:focal} as:

\begin{equation}
\begin{aligned}
\label{eq:new_loss}
    \mathcal{L}_{cls}^{\text{(new)}} = & \sum_{i=1}^{N_{pos}}{(|\textcolor{red}{f_{1}(s_i)}-p_i|^{\gamma} \mathrm{BCE}(p_i, \textcolor{red}{f_{1}(s_i)})}\\
    & + \sum_{i=1}^{N_{neg}} p_i^{\gamma} \mathrm{BCE}(p_i, 0),
\end{aligned}
\end{equation}

\noindent
where we mark the difference with Eq. \ref{eq:focal} in \textcolor{red}{red}. We use the $s_i$ as a positional metric like IOU between the $i$\textsuperscript{th} ground truth and its corresponding prediction. As some examples, we can use $f_1(s_i)$ as $s_i$, $s_i^2$, and $e^{s_i}$ in implementations.

In our experiments, We found that $f_1(s_i)=\varepsilon(s_i^2)$ works best in our implementations, where $\varepsilon$ is a transformation to rescale numbers to avoid some degenerated solutions, as $IOU$ values may be very small sometimes. We tried two rescale strategies, first is to ensure the highest $s_i^2$ is equal to the max $IOU$ value among all possible pairs in a training example, which is inspired by \cite{TOOD}, and the other is to ensure the highest $s_i^2$ is equal to $1.0$, which is a simpler way. We find the former works better for detectors with more queries like DINO ($900$ queries), and the latter works better for detectors with $300$ queries.

The design tries to supervise classification scores with positional metrics like IOU. It encourages predictions with low classification scores and high IOU scores, while penalizing predictions with high classification scores but low IOU scores. 

\subsection{Position-Modulated Matching}
\label{sec:IOU_matching}

The position-supervised classification loss aims to encourage predictions with high IOU scores but low classification scores. Following the spirit of the new loss, we would like to make some modifications to the matching costs. We rewrite Eq. \ref{eq:cls_cost} as follows:

\begin{equation}
\begin{aligned}
\mathcal{C}_{cls}^{\text{(new)}}(i, j) = & {|1-p_i \textcolor{red}{f_2(s_i')} |^{\gamma} \mathrm{BCE}(p_i\textcolor{red}{f_2(s_i')} , 1)} \\
    &- (p_i \textcolor{red}{f_2(s_i')}) ^{\gamma} \mathrm{BCE}(1 - p_i \textcolor{red}{f_2(s_i')} , 1), \\
\end{aligned}
\end{equation}

\noindent
where we mark the difference with Eq. \ref{eq:cls_cost} in \textcolor{red}{red}. $s_i'$ is another positional metric, which we use a rescaled GIOU in our implementations. As GIOU ranges from [-1,1], we shift and rescale it to the range [0,1] as a new metric. $f_2$ is another function to tune. We empirically use $f_2(s_i') = (s_i')^{0.5}$ in our implementations. 

Intuitively, the $f_2(s_i')$ is used as a modulated function to down-weight the predictions with inaccurate prediction boxes. It helps to align classification scores and bounding box predictions better as well.

One interesting question is why we do not directly use the new classification loss (Eq. \ref{eq:new_loss}) as a new classification cost. The matching is calculated between all predictions and ground truths, under which there will be many low-quality predictions. Ideally, we hope a prediction with a high IOU score and a high classification score will be selected as a positive example for its low matching cost. However, a prediction with a low IOU score and a low classification score will also have a low matching cost, making the model degenerative.

\subsection{Analyses}
\label{sec:analysis}

\subsubsection{Why Supervise Classification with Positional Scores only?}

We argue that the source of unstable matching is the \textit{multi-optimization path problem}. Discuss the simplest scenario: We have two imperfect predictions, A and B. 
As shown in Fig. \ref{fig:multi_optimization_path}, prediction A has a higher IOU score, but a lower classification score since its center locates in the background. In contrast, prediction B has a larger classification score but a lower IOU score. The two predictions will compete for the ground truth object. If anyone is assigned a positive example, the other will be set as a negative one. A ground truth with two imperfect candidates is common during training, especially in the early steps.

Due to the randomness during training, Each one of the two predictions has a probability of being assigned as a positive example. Under the default DETR variants loss designs, each possibility will be amplified since the default loss design will encourage positive and restrain negative examples, as shown in table \ref{table:multi_opt_path}. Detection models have two different optimization paths: models prefer high IOU samples or high classification score samples. The different optimization paths can confuse the model during training. A good question is if the model can encourage both predictions. Unfortunately, it will violate the requirements of one-to-one matching. The problem is not significant in traditional detectors, which assign multiple predictions to each ground truth. The one-to-one matching strategy in DETR-like models will amplify the conflicts. 

In contrast, if we supervise classification scores with positional metrics (like IOU), the problem will be eliminated, as shown in the last row of Table \ref{table:multi_opt_path}. Only Prediction A will be encouraged toward the target. If prediction B is matched, it will not be optimized continuously since it has a low IOU score. There will be only one optimization path for the model, which will stable the training. 

How about using classification information to supervise classification scores? 
Some previous work in traditional detectors tried to align classification and IOU scores by using a quality score \cite{TOOD, GFL}, which is a combination of both classification and IOU scores. 
Unluckily, the design is not suitable for DETR-like models, which will be shown in Sec. \ref{sec:abla_loss_design}, as it cannot solve the root of the unstable matching, \textit{multi-optimization path problem}.
Suppose both classification and IOU scores are included in the targets. In that case, prediction B will also be encouraged if matched since it has a high classification score. The \textit{multi-optimization path problem} also exists, which damages the model training.

\input{tables_tex/multi_optimization_path}

Another direct question is whether we can optimize the model toward another path. If we would like to guide models to prefer a high classification score, i.e., encourage matching prediction B in the example. There will be ambiguity if there are two objects of the same category. For example, there are two cats in an image. The classification score is determined by semantic information, which means that a box near any cat will have a high classification score, which can damage the model training.

\input{images_tex/unstable_score}

\subsubsection{Rethink the Role of Classification Scores in Detection Transformers}

The new matching loss connects the DETR-like models to traditional detectors as well. Our new loss design shares a similar optimization path as traditional detectors.

An object detector has two optimization paths: one is to find a good predicted box and optimize its classification score; the other is to optimize a prediction with a high classification score to the ground truth box. Most traditional detectors assign predictions by checking their positional accuracy only. The models encourage anchor boxes that are near the ground truth. It means that most traditional detectors select the first optimization way.  
Differently, DETR-like matching additionally considers classification scores and uses the weighted sum of classification and localization scores as the final cost matrix. The new matching way results in conflicts between the two ways. 

Since then, why still DETR-like models used classification scores during Training? We argue that it is more like a reluctant design for one-to-one matching. Previous work \cite{PeizeSun2020WhatMF} has shown that introducing classification cost is the key to one-to-one matching. It can ensure only one positive example of the ground truth. As the localization losses (box L1 loss and GIOU loss) do not restrain negative examples, all predictions near a ground truth will be optimized toward the ground truth. 
There will be unstable results if only position information is considered during matching. With the classification scores in the matching, the classification scores are \textit{used as marks to denote which prediction should be used as positive examples}, which can promise a stable matching during training compared with position-only matching.

However, as the classification scores are optimized independently, without any interaction with positional information, it sometimes leads the model to another optimization path, i.e., encourage the box with a larger classification score but a worse IOU score. Our position-supervised loss can help to align the classification and localization, which not only ensures a one-to-one matching, but also solves the multi-optimization problem.

With our new loss, the DETR-like models work more like traditional detectors as they both encourage predictions with larger IOU scores but a worse classification score.

\subsubsection{Comparisons of Unstable Scores}
To present the effectiveness of our methods. We compare the unstable scores between vanilla DINO and DINO with stable matching in Fig. \ref{fig:unstable_score}.  
The unstable scores are the inconsistent matching results between adjacent decoder layers. For example, if we have $10$ ground truth boxes in an image, and only one box has a different prediction indexed matched in the $(i-1)$\textsuperscript{th} and $i$\textsuperscript{th} layers, then the unstable score of the layer $i$ is $1/10 \times 100.00 = 10.00 \%$.
Typically, a model has six decoder layers. The unstable score of layer $1$ is calculated by comparing the matching results of the encoder and the first decoder layer.

We use model checkpoints at the 5000th step and evaluate models on the first 20 images in the COCO \texttt{val2017} dataset. The results show that our model is more stable than DINO. There are two interesting observations in Fig. \ref{fig:unstable_score}. First, the unstable score generally decreases from the first decoder layer to the last decoder layer, which means the higher decoder layers (with larger indexes) may have more stable predictions. Moreover, there is a strange peak in layer $5$ of DINO's unstable scores, while DINO with stable matching does not. We suspect some randomness causes the peak.

\section{Memory Fusion}
\label{sec:memry_fusion}

\input{images_tex/feature_fusion}

To further enhance the model convergence speed at the early training stage. We proposed a straightforward feature fusion technique termed memory fusion, which involves merging the encoder output features at different levels with the multi-scale backbone features. 
We propose three different memory fusion ways, named simple fusion, U-like fusion, and dense fusion, which are shown in Fig. \ref{fig:feature_fusion} (b), (c), and (d). 
For multiple features to fuse, we first concatenate them along the feature dimension and then project the concatenated feature to the original dimensions. More implementation details of memory fusion are available in Appendix Sec. \ref{sec:detail_memory_fusion}.

The dense fusion achieves better performance in our experiments, which is used as our default feature fusion. We compare the training curves of DINO and DINO with dense fusion in Fig.\ref{fig:convergence_speed_with_matching_and_fusion}. It shows that the fusion enables a faster convergence, especially in the early steps.

\input{images_tex/compare_dino_stabledino_stablefusion}

\input{tables_tex/COCO_R50}
\input{tables_tex/COCO_SwinL}
\input{tables_tex/MaskDINO_COCO}

\section{Experiments}
\vspace{-0.05cm}
\subsection{Settings}
\vspace{-0.05cm}
\paragraph{Dataset.} We conduct experiments on the COCO 2017 object detection dataset~\cite{lin2014microsoft}. All models were trained using the \texttt{train2017} set without extra data and evaluated their performance on the \texttt{val2017} set. We report our results with two different backbones, including ResNet-50~\cite{He_2016_CVPR} pretrained on ImageNet-1k~\cite{imagenet} and Swin-L~\cite{liu2021swin} pretrained on ImageNet-22k~\cite{imagenet}.
\vspace{-0.4cm}

\paragraph{Implementation Details.} We test the effectiveness of our stable matching strategy based on DINO~\cite{zhang2022dino}. We trained our models on COCO training using AdamW optimizer~\cite{loshchilov2017fixing, DiederikPKingma2014AdamAM} with a learning rate of $1 \times 10^{-4}$ for 12 epochs, and the learning rate is reduced by a factor of 0.1 at the 11\textsuperscript{th} epoch. In the case of 24-epoch settings, the learning rate is decreased at the 20\textsuperscript{th} epoch. We set the weight decay to $10^{-4}$. 
We conduct all our experiments based on detrex \cite{ideacvr2022detrex}. We follow their hyperparameters as default for other DETR variants. As the new loss design results in a smaller scale of classification loss, we empirically choose a $6.0$ as the classification loss weight.
Moreover, we found a proper Non-Maximum Suppression (NMS) can still help the final performance for about $0.1-0.2$ AP. We use NMS by default with a threshold $0.8$.
We use a random seed 60 in all our experiments to ensure the results are reproducible. DINO with the seed $60$ in detrex \cite{ideacvr2022detrex} has the same results (49.0 AP) as the original paper.

\subsection{Main Results}
As shown in Table \ref{tab:compare_with_r50_backbone}, we firstly compare our Stable-DINO on COCO object detection \texttt{val2017} set with other DETR variants with ResNet-50~\cite{he2015deep} backbone. Stable-DINO-4scale and Stable-DINO-5scale can achieve 50.2 AP and 50.5 AP on $1 \times$ scheduler, which gains $1.2$ and $1.1$ AP over the DINO-4\&5 scale $1 \times$ baselines. And with $2 \times$ training scheduler, Stable-DINO-4scale even increased AP by $1.1$ and $0.6$ compared with DINO-4scale $2 \times$ and $3 \times$ baselines. 
Table \ref{tab:compare_with_swin_backbone} compares our models to other state-of-the-art Transformer-based detectors with large backbones, such as the ImageNet-22k~\cite{imagenet} pre-trained Swin-Large backbone. Stable-DINO-4scale can achieve 57.7 AP on $1 \times$ and obtain 58.6 AP for $2 \times$ scheduler, which outperforms the DINO $1 \times$ and $3 \times$ baselines by 0.9 and 0.6 AP.

Comparison with SOTA methods are available in Table \ref{tab:sota}.

\subsection{Generalization of our Methods}
To verify the generalization of our models, we ran experiments on other DETR variants. The results are available in Table \ref{table:generalization}. Our methods show consistent improvement on existing models, including Deformable-DETR \cite{zhu2020deformable}, DAB-Defomable-DETR \cite{liu2022dabdetr}, and  $\mathcal{H}$-DETR \cite{chen2019hybrid}. 

\input{tables_tex/generalization}

To further present the effectiveness of our methods on different tasks, we implement our methods on MaskDINO \cite{li2022mask} for both object detection and segmentation. We name the new model Stable-MaskDINO. 
Stable-MaskDINO outperforms MaskDINO on both detection and segmentation tasks, as shown in Fig. \ref{tab:maskdino}.


\subsection{Ablation Study}

We present ablations in this section. We use ResNet-50 backbones and 12-epoch training as the default setting.
\vspace{-0.5cm}

\paragraph{Effectiveness of model designs.}
We first verify the effectiveness of each design in our model. The results are available in Table \ref{tab:ablation}. To make a fair comparison, we test DINO with NMS $0.8$ in row $1$ of the table. The model has $0.2$ gains compared with the default test way,

The results show that the position-supervised loss and the position-modulated cost help the final results, with +0.6 AP and +0.4 AP gains, respectively. It is worth noting that the DINO has achieved high performance already; hence each gain is hard to obtain.  

We find that dense fusion works best among the three ways by comparing the different memory fusion ways. It brings $+0.2$ AP and $+0.5$ AP$_{50}$ compared with baselines. Moreover, the fusion helps a lot during the early training steps, as shown in Sec. \ref{sec:memry_fusion}.


\vspace{-0.5cm}
\paragraph{Comparisons of different loss designs.}
\label{sec:abla_loss_design}

We compared the effectiveness of different loss designs in Table \ref{table:loss_design}. We ignore the transformation function $\varepsilon$ (see Sec. \ref{sec:IOU_loss}) for simplifications in the table. To keep a fair comparison, we use loss weight as $10.0$ in all experiments in the table. All models train trained without memory fusion and position-modulated cost.

There are some interesting observations in the experiments. First, with positional metrics as supervision, the model has performance gains most of the time. The methods are robust to the function design. For example, it even works well with the $f_1(s)=(e^s-1)/(e-1)$ function. Second, the introduction of classification scores (like probabilities) will result in a performance drop in models, as shown in the lines $5$, $6$, and $7$ in Table \ref{table:loss_design}. it verifies our analysis in Sec. \ref{sec:intro} and Sec. \ref{sec:analysis}. It also demonstrates the effectiveness of our methods design. At last, convex functions like $f_1(s)=s^2$ work better than concave functions like $f_1(s)=s^{0.5}$. As a special case, the concave function $\sin(s \times \pi / 2)$ even results in a performance drop, since it reaches $1$ fast with the increasing of $s$. 

\vspace{-0.5cm}
\paragraph{Comparisons of different loss weights.}
\label{sec:abla_loss_weight}

We test different loss weights for our position-supervised loss in this section. The results are available in Table \ref{table:loss_weight}. The results show that our model works well for most classification weights, e.g. from $4.0$ to $10.0$. We use the position-modulated cost and use no memory in this ablation.

\vspace{-0.5cm}
\paragraph{Ablations for the position-modulated cost.}
\label{sec:abla_cost}

We compare results with different function and cost weight designs in this section. The results are available in Table \ref{table:abla_cost}. We choose $f_2(s)=s^{0.5}$ and cost weight $2.0$ by default.

\input{tables_tex/abltion}
\input{tables_tex/loss_design}
\input{tables_tex/loss_weight}
\input{tables_tex/ablation_cost}

\section{Related Work}
\paragraph{Detection Transformer.}
Detection Transformer (DETR) \cite{carion2020end} proposed a new detector with Transformer-based heads and eliminated the dependents of head-craft modules. Although the novelty design, it suffers from slow convergence and inferior performance. Many follow-ups try to solve the problem from different perspectives. For example, some work \cite{gao2021fast, meng2021conditional, anchordetr, liu2022dabdetr} found the importance of positional priors and propose to add more positional prior to models. DAB-DETR \cite{liu2022dabdetr}, as an example, formulated the decoder queries as dynamic anchor boxes for better results.
Some work \cite{zhu2020deformable, nguyen2021boxer} designed new operators to fasten model training, like deformable attention in Deformable DETR.
Another line of work \cite{li2022dn, HDETR, groupdetr, codetr2022} tried to add extra branches to the decoder. They found that auxiliary tasks can help the convergence of models. 
There are other explorations with traditional matching \cite{ouyang2022nms}, model pre-train \cite{dai2021up}, and so on. 

Although their great progress, the unstable matching problem across decoder layers got less attention. 
We analyze the reason for the unstable matching problem and propose a simple but priceless solution to the problem in this paper. The new loss and matching designs introduce marginal costs to previous work, resulting in better model performance.

\vspace{-0.5cm}
\paragraph{Variants of Focal Loss.}
Our loss design is a variant of Focal loss \cite{lin2017focal}. Although it is less focused on DETR variants, there are many works \cite{TOOD, GFL, NabilaAbraham2018ANF} that focus on loss improvement in classical detectors. The most related work to our solution is the task-aligned loss \cite{TOOD}. We have a different motivation with the task-aligned loss. We focus on the stable matching problem in DETR variants, which does not exist in traditional detectors. 
Moreover, although the great results of task-aligned loss in one-stage detectors\footnote{One-stage and two-stage detectors are commonly used concepts in the classical detectors. We view DETR variants as a new design and do not classify them as the one-stage or two-stage detectors.}, the loss cannot be used to DETR variants directly. It introduces classification scores as extra supervision, which results in performance drops in the DETR-like models with one-to-one matching, as shown in Sec. \ref{table:loss_design}. The key reason for the difference is the two matching ways between traditional detectors and our solutions. 

In our paper, we first analyze the phenomenon of unstable matching and point out that the key is the multi-optimization path problem. Then we show that the most crucial design to solve the problem is to use and only use positional metrics to supervise classification scores. We provide a neater and more principal solution to the unstable matching problem in DETR-like models.

\section{Conclusion}

We analyze the stable matching in DETR-like models and present that the root cause of the problem is the multi-optimization path problem. To solve the multi-optimization path problem, we present that the key to solving it is to use positional metrics to supervise classification scores of positive examples. We then propose a new position-supervised loss and a new position-modulated cost for DETR-like models. Moreover, we propose a dense memory fusion to enhance encoder and backbone features. We validate the effectiveness of our design on many DETR-like variants.

\noindent
\textbf{Limitations.} Although our method shows great performance, we only verify it on DETR-like image object detection and segmentation. More explorations like 3D object detection will be left as our future work. Moreover, we focus on the classification parts in the loss and matching only, while the localization parts are preserved. Analyses for the localization parts are left as our future work as well.


\clearpage
{\small
\bibliographystyle{ieee_fullname}
\bibliography{egbib}
}

\clearpage
\appendix

\section{More Experiments about Stable Matching}
\label{sec:more_stable}

We visualize the queries with top 30 IOU scores of DINO and Stable-DINO in Fig. \ref{fig:compare_top30}. It shows that Stable-DINO has better alignment between IOU and probability scores.

\input{images_tex/compare_top30}

\section{Details of Memory Fusion}
\label{sec:detail_memory_fusion}

\paragraph{The implementation of memory fusion.} The implementation of memory fusion has been depicted in Fig.\ref{fig:memory_fusion}.  The fusion was executed in a very simple way.  The outputs from each encoder layer were amassed and subsequently concatenated with the backbone features along the channel dimension. Following concatenation, a linear projection layer, in conjunction with a norm layer, was employed to project the channel dimension, aligning it to the dimension of the decoder layer. And the fused features were then forwarded to the decoding stage.

\paragraph{How Memory Fusion Works?} In the DETR variants, a pre-trained backbone model is often utilized for feature extraction from the input raw images which is typically pre-trained on large-scale dataset such as ImageNet~\cite{imagenet}. The extracted features are merged with position encodings and fed into the transformer encoder for extracting and fusing global and local information. While the encoder and backbone can be seen as the same meta-framework for feature extraction but differ in their initialization ways. The encoder's weights are randomly initialized, whereas the backbone features are pre-trained, which means the encoder's feature extraction capability is insufficient in the early stages of training. By fusing the pre-trained backbone features with the multi-scale features processed by the encoder, we enable the decoder to better utilize the pre-trained backbone features during the early training stages. As illustrated in Fig.\ref{fig:convergence_speed_with_matching_and_fusion}, our stable matching strategy significantly accelerates the convergence speed in the early iterations of the training process. Moreover, our newly designed dense memory fusion technique can further boost the convergence speed based on this foundation.

\input{images_tex/memory_fusion}

\section{SOTA experiments}
To verify the scalability of our models, we verify our Stable-DINO with large-scale datasets and models. After pre-trained on Objects365 \cite{shao2019objects365}, Stable-DINO reaches $63.7$ AP on \texttt{val2017} and $63.8$ AP on \texttt{test-dev} without test-time augmentation. We set a new SOTA with under the same setting. The results are available in Table \ref{tab:sota}.

\begin{table*}[h]
    \centering
        \footnotesize
            \renewcommand{\arraystretch}{1.3}
    \resizebox{1.0\textwidth}{!}{%
    \begin{tabular}{c|c|c|c|c|c|c|c|c}
        \shline
        {Method} & {Params} & Backbone Pre-training Dataset  & Detection Pre-training Dataset & Use Mask & Use TTA & End-to-end  &  val2017 (AP) & test-dev (AP) \\
        \shline
        SwinL~\cite{liu2021swin} & $284$M & IN-22K-14M & O365  & \checkmark & \checkmark & & $58.0$ & $58.7$ \\
        DyHead~\cite{dai2021dynamic} & $\geq 284$M & IN-22K-14M & Unknown*  &  & \checkmark &   & $58.4$  & $60.6$ \\
        Soft Teacher+SwinL~\cite{xu2021end} & $284$M & IN-22K-14M & O365  & \checkmark & \checkmark & & $60.7$  & $61.3$ \\
        GLIP~\cite{li2021grounded} & $\geq 284$M & IN-22K-14M & FourODs~\cite{li2021grounded},GoldG+~\cite{li2021grounded,kamath2021mdetr} &  &\checkmark& & $60.8$ & $61.5$\\
        Florence-CoSwin-H\cite{LuYuan2022FlorenceAN} & $\geq 637$M & FLD-900M~\cite{LuYuan2022FlorenceAN} &  FLD-9M~\cite{LuYuan2022FlorenceAN} &  &\checkmark&  & $62.0$ & $62.4$ \\
        SwinV2-G~\cite{liu2021swinv2} & $3.0$B & IN-22K-ext-70M~\cite{liu2021swinv2} & O365 & \checkmark &\checkmark&  & $62.5$ & $63.1$\\
        DINO-SwinL & ${218}$\textbf{M} & IN-22K-14M & O365 & & \checkmark& \checkmark& $\textbf{63.2}$ & $\textbf{63.3}$ \\
        \hline
        Stable-DINO-SwinL(Ours) & $\textbf{218}$\textbf{M} & IN-22K-14M & O365 &  & & \checkmark& $\textbf{63.7}$ & $\textbf{63.8}$ \\
        \shline
    \end{tabular}
    }
    \vspace{0.05cm}
    \caption{Comparison of the best detection models on MS-COCO. Similar to DETR~\cite{carion2020end}, we use the term ``end-to-end'' to indicate if a model is free from hand-crafted components like RPN and NMS. The term ``TTA'' means test-time augmentation. The term ``use mask'' means whether a model is trained with instance segmentation annotations. We use the terms ``IN'' and ``O365'' to denote the ImageNet~\cite{deng2009imagenet} and Objects365~\cite{shao2019objects365} datasets, respectively. Note that ``O365'' is a subset of ``FourODs'' and ``FLD-9M''. * DyHead does not disclose the details of the datasets used for model pre-training.
    }
    \label{tab:sota}
\end{table*}

\section{Convergence Comparison}
We compare the convergence speed of Satble-DINO and DINO in Fig. \ref{fig:convergence_dino_stabledino}. It shows that Stable-DINO convergence faster than DINO. 

\input{images_tex/compare_dino_stabledino_convergence}

\section{Visualizations of Encoder Sampling Points }
To understand the effectiveness of memory fusion, we compare the sampling points of DINO and DINO with memory fusion in Fig. \ref{fig:compare_sampling_points}. We use the first checkpoint, i.e., 5000 iteration steps, during training for the visualization. As DINO use deformable attention in the Transformer encoder, we plot reference points (blue stars) and corresponding sampling points (red to yellow dots) in the figure.
The results show that the memory fusion enable models to cover long range features, which introduces more global information for the model. We suspect that the encoder need to fuse global information to the features from backbones. The residual connections among encoder layers in the memory fusion accelerates the process.

\input{images_tex/comapre_sampling_points}

\section{Visualizations of Model Results}
We visualize the results of DINO and Stable-DINO in Fig. \ref{fig:vis_1}. Stable-DINO has better results than DINO for its more accuracy predictions. For example, DINO has an wrong prediction ``car'' as shown in the first row of Fig. \ref{fig:vis_1}. Similarly, DINO marks the sky as a ``kite'' in the last row of Fig. \ref{fig:vis_1}, while Stable-DINO does not. It does not mean that Stable-DINO is more conservative, as it predicts the bicycles in the third row of Fig. \ref{fig:vis_1}. Stable-DINO has better visualization results compared with DINO.

\input{images_tex/vis_1}

\end{document}

%% file: images_tex/performance_compare.tex
\begin{figure}[t]
    \centering
    \includegraphics[width=1.0\linewidth]{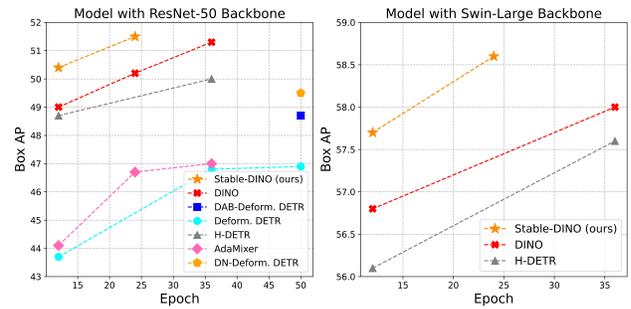}
    \vspace{-0.7cm}
    \caption{Comparison of our methods (named Stable-DINO in figures) and baselines. We compare models with ResNet-50 backbones in the left figure and models with Swin-Transformer Large backbones in the right figure. All models use a maximum $1/8$ resolution feature map from a backbone, except AdaMixer uses a maximum $1/4$ resolution feature map.
    }
    \label{fig:model_comparison}
    \vspace{-0.4cm}
\end{figure}

%% file: images_tex/multi_optimization_path.tex
\begin{figure}[t]
    \centering
    \includegraphics[width=1.0\linewidth]{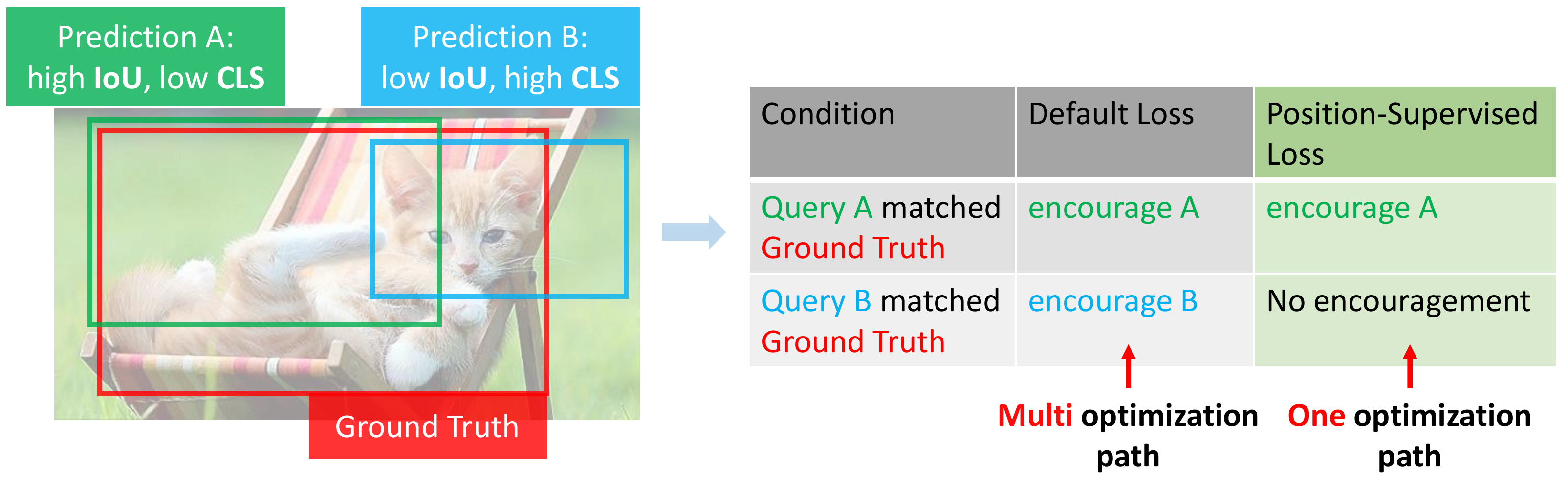}
    \vspace{-0.6cm}
    \caption{Explanation of the \textit{multi-optimization path} problem. We use the term ``CLS'' as classification scores. 
    Each prediction has a probability to be assigned as the positive example in bipartite matching and be encouraged towards ground truth during training, which can be different optimization paths. With the position-supervised loss, only one optimization path will have in the training, which can stabilize the matching. 
    }
    \label{fig:multi_optimization_path}
    \vspace{-0.5cm}
\end{figure}

%% file: tables_tex/multi_optimization_path.tex
\begin{table}[h]
\begin{center}
\resizebox{0.85\columnwidth}{!}{%
\begin{tabu}{
l@{\hskip9pt} |  
l@{\hskip9pt} |  
l@{\hskip9pt} 
}
 \toprule
 & 
 \green{Prediction A} Matched & 
 \blue{Prediction B} Matched 
\\
\midrule
   \multirow{2}{*}{Default Matching} & encourage A & restrain A \\
                         & restrain B & encourage B \\

\midrule
   \multirow{2}{*}{Stable Matching} & encourage A & restrain A slightly \\
                                & restrain B & No encourage \\
  \bottomrule
\end{tabu}}
\vspace{-0.2cm}
\caption{Detailed explanation of the multi-optimization path problem. Suppose we have two imperfect predictions: A with a higher IOU score and lower classification score, while B is on the opposite. An example is shown in Fig.\ref{fig:multi_optimization_path}.}
\label{table:multi_opt_path}
\vspace{-0.2cm}
\end{center}
\end{table}

%% file: images_tex/unstable_score.tex
\begin{figure}[t]
    \centering
    \includegraphics[width=0.9\linewidth]{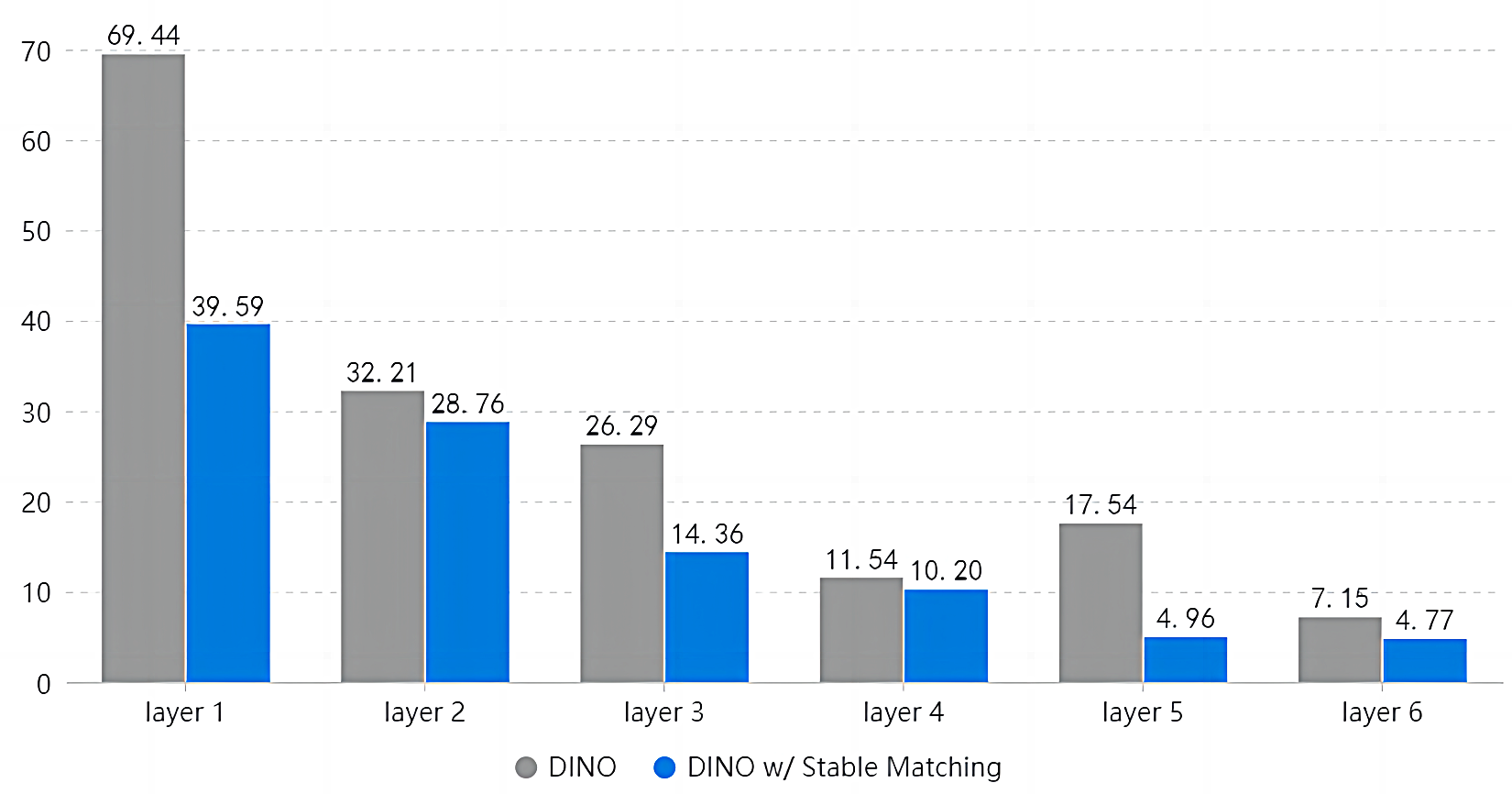}
    \vspace{-0.4cm}
    \caption{Comparisons of the unstable scores of DINO and DINO with stable matching. 
    }
    \label{fig:unstable_score}
    \vspace{-0.4cm}
\end{figure}

%% file: images_tex/feature_fusion.tex
\begin{figure}[t]
    \centering
    \includegraphics[width=1.0\linewidth]{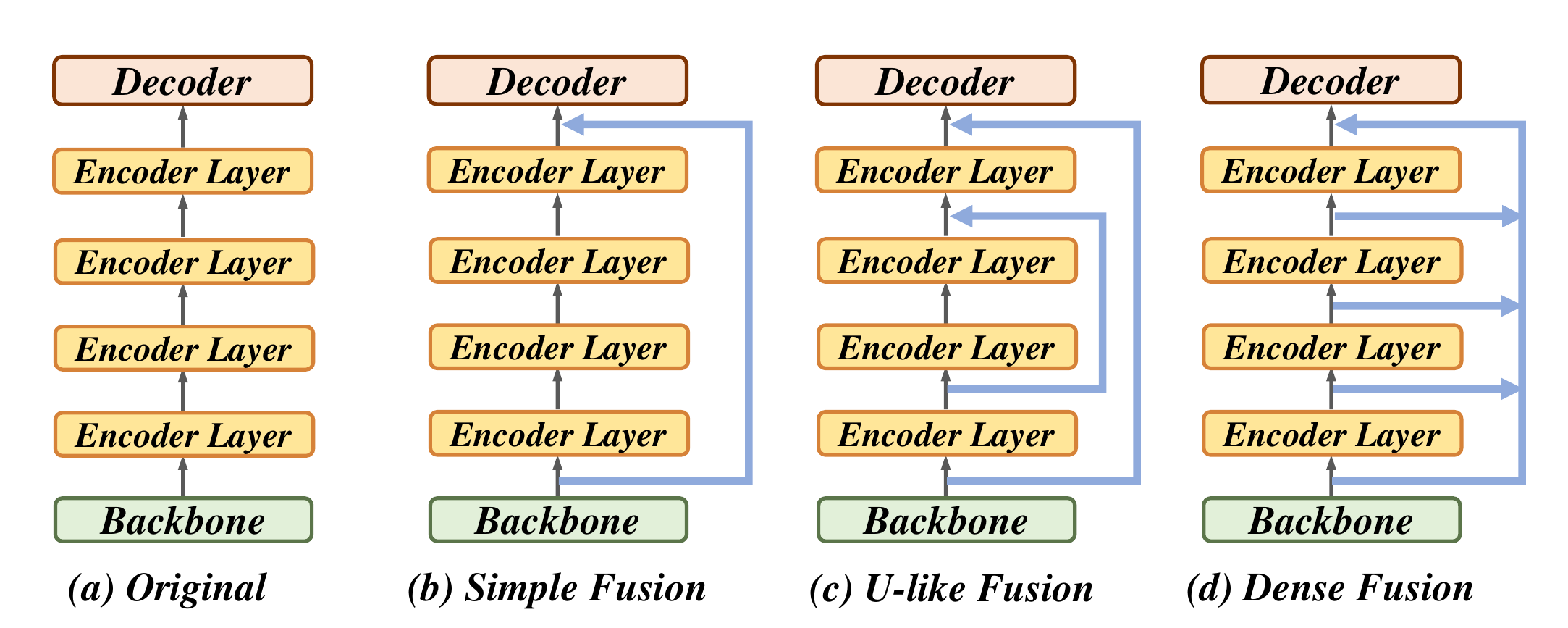}
    \vspace{-0.8cm}
    \caption{Comparison of our methods and baselines. We compare the (a) original memory feature with our proposed three memory fusion ways: (b) simple memory fusion, (c) U-like memory fusion, and (d) dense memory fusion.
    }
    \label{fig:feature_fusion}
    \vspace{-0.4cm}
\end{figure}

%% file: images_tex/compare_dino_stabledino_stablefusion.tex
\begin{figure}[t]
    \centering
    \includegraphics[width=0.7\linewidth]{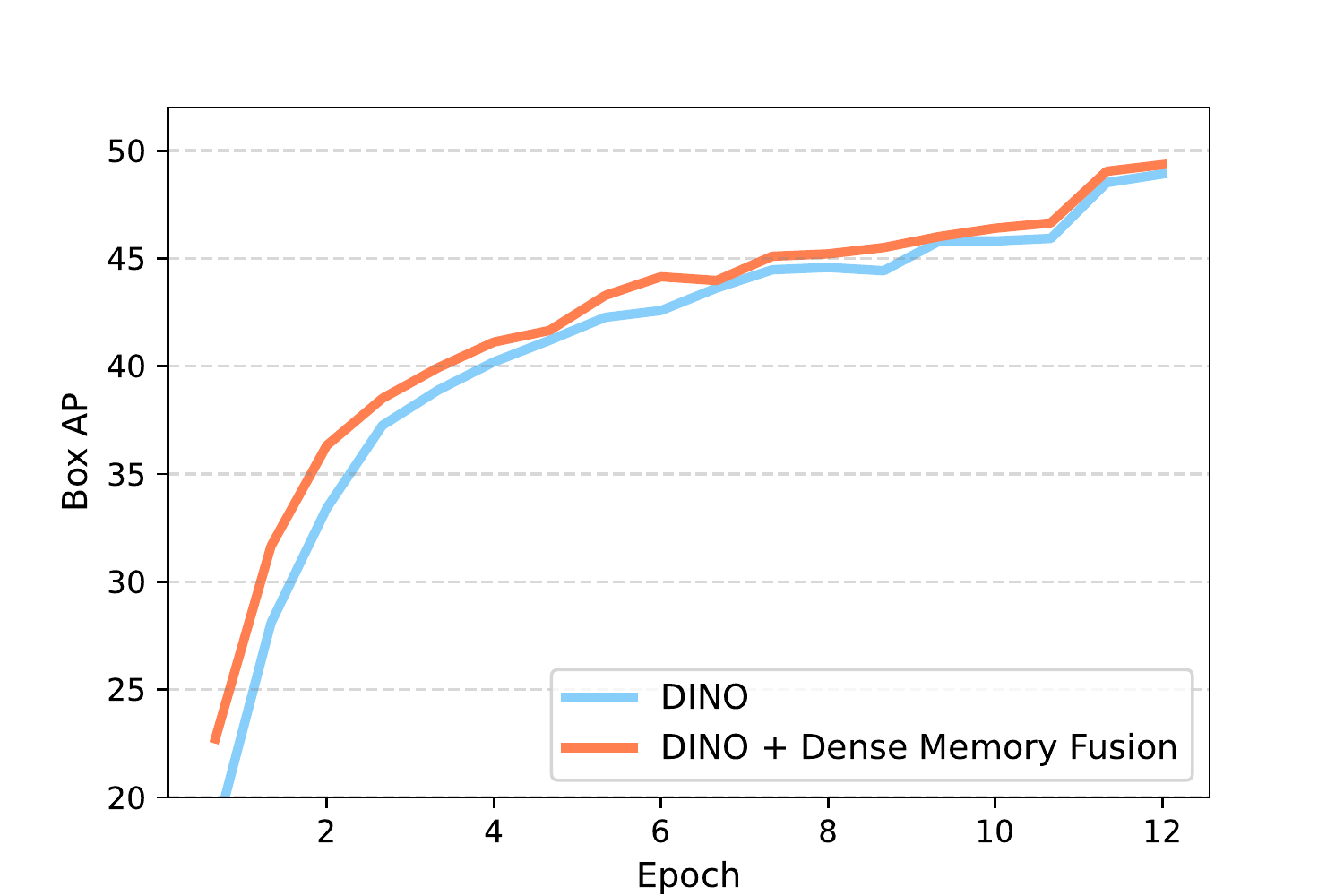}
    \vspace{-0.3cm}
    \caption{Comparison of the convergence speed of DINO and DINO with our dense memory fusion.
    }
    \label{fig:convergence_speed_with_matching_and_fusion}
    \vspace{-0.4cm}
\end{figure}

%% file: tables_tex/COCO_R50.tex
\begin{table*}[ht]
\centering\setlength{\tabcolsep}{7pt}
\renewcommand{\arraystretch}{1.2}
\footnotesize
\vspace{-3mm}
{
\resizebox{0.8\textwidth}{!}{%
\begin{tabular}{l|l|c|cccccc}
\shline
Model & Backbone & \#epochs  & AP & AP$_{50}$ & AP$_{75}$ & AP$_{S}$ & AP$_{M}$ & AP$_{L}$\\
\shline
Conditional-DETR~\cite{meng2021conditional} & R$50$  & $108$ & $43.0$ & $64.0$ & $45.7$ & $22.7$ & $46.7$ & $61.5$ \\
SAM-DETR~\cite{zhang2022SAMDETR} & R$50$  & $50$ & $39.8$ & $61.8$ & $41.6$ & $20.5$ & $43.4$ & $59.6$ \\
SAM-DETR + SMCA~\cite{zhang2022SAMDETR} & R$50$  & $50$ & $41.8$ & $63.2$ & $43.9$ & $22.1$ & $45.9$ & $60.9$ \\
Anchor-DETR~\cite{anchordetr} & R$50$  & $50$ & $42.1$ & $63.1$ & $44.9$ & $22.3$ & $46.2$ & $60.0$ \\
Dynamic-DETR~\cite{dynamicdetr} & R$50$  & $12$ & $42.9$ & $61.0$ & $46.3$ & $24.6$ & $44.9$ & $54.4$ \\
SMCA-DETR~\cite{gao2021fast} & R$50$  & $108$ & $45.6$ & $65.5$ & $49.1$ & $25.9$ & $49.3$ & $62.6$ \\
AdaMixer~\cite{adamixer22cvpr} & R$50$  & $36$ & $47.0$ & $66.0$ & $51.1$ & $30.1$ & $50.2$ & $61.8$ \\
CF-DETR~\cite{cfdetr2022} & R$50$  & $36$ & $47.8$ & $66.5$ & $52.4$ & $31.2$ & $50.6$ & $62.8$ \\
Sparse-DETR~\cite{roh2021sparse} & R$50$ & $50$ & $46.3$ &  $66.0$ & $50.1$ & $29.0$ & $49.5$ & $60.8$ \\
Efficient-DETR~\cite{yao2021efficient} & R$50$ & $36$ & $45.1$ & $63.1$ & $49.1$ & $28.3$ &  $48.4$ & $59.0$ \\
BoxeR-2D~\cite{nguyen2021boxer} & R$50$ & $50$ & $50.0$ & $67.9$ & $54.7$ & $30.9$ &  $52.8$ & $62.6$ \\
Deformable-DETR~\cite{zhu2020deformable}  & R$50$   &$50$ & $46.2$ &  $65.0$ & $50.0$ & $28.3$ & $49.2$ & $61.5$  \\
Deformable-DETR~\cite{zhu2020deformable}  & R$50$  &$50$ & $46.9$ & $65.6$ & $51.0$ & $29.6$ &  $50.1$ & $61.6$ \\
DAB-Deformable-DETR~\cite{liu2022dabdetr} & R$50$  & $50$ & $46.8$ & $66.0$ & $50.4$ & $29.1$ & $49.8$ & $62.3$ \\
DN-Deformable-DETR~\cite{li2022dn}  & R$50$  & $12$ & $43.4$ & $61.9$ & $47.2$ & $24.8$ & $46.8$ & $59.4$ \\
DN-Deformable-DETR~\cite{li2022dn}  & R$50$ &$50$ & $48.6$ & $67.4$ & $52.7$ & $31.0$ & $52.0$ & $63.7$ \\ 
$\mathcal{H}$-DETR~\cite{HDETR} & R$50$  & $12$ & $48.7$ & $66.4$ & $52.9$ & $31.2$ & $51.5$ & $63.5$ \\
$\mathcal{H}$-DETR~\cite{HDETR} & R$50$  & $36$ & $50.0$ & $68.3$ & $54.4$ & $32.9$ & $52.7$ & $65.3$ \\
Co-DETR~\cite{codetr2022} & R$50$  & $12$ & $49.5$ & $67.6$ & $54.3$ & $32.4$ & $52.7$ & $63.7$ \\
DINO-4scale~\cite{zhang2022dino} & R$50$  & $12$ & $49.0$ & $66.6$ & $53.5$ & $32.0$ & $52.3$ & $63.0$ \\
DINO-5scale~\cite{zhang2022dino} & R$50$ & $12$ & $49.4$ & $66.9$ & $53.8$ & $32.3$ & $52.5$ & $63.9$ \\
DINO-4scale~\cite{zhang2022dino} & R$50$  & $24$ & $50.4$ & $68.3$ & $54.8$ & $33.3$ & $53.7$ & $64.8$ \\ 
DINO-4scale~\cite{zhang2022dino} & R$50$  & $36$ & $50.9$ & $\mathbf{69.0}$ & $55.3$ & $34.6$ & $54.1$ & $64.6$ \\
\shline
\rowcolor{gray!15} Stable-DINO-4scale (ours) & R$50$  & $12$ & $50.4$ {(+$1.4$)} & $67.4$ & $55.0$ & $32.9$ & $54.0$ & $65.5$ \\
\rowcolor{gray!15} Stable-DINO-5scale (ours) & R$50$  & $12$ & $50.5$ {(+$1.1$)} & $66.8$ & $55.3$ & $32.6$ & $54.0$ & $65.3$ \\
\rowcolor{gray!15} Stable-DINO-4scale (ours) & R$50$ & $24$ & {$\mathbf{51.5}$} {(+$1.1$)} & ${68.5}$ & $\mathbf{56.3}$ & $\mathbf{35.2}$ & $\mathbf{54.7}$ & $\mathbf{66.5}$ \\
\shline
\end{tabular}}
\vspace{-0.2cm}
\caption{Comparison to prior DETR variants on COCO \texttt{val2017} with ResNet-50 backbones. The numbers in brackets are AP improvements compared with corresponding DINO models under the same settings.}
\vspace{0.2cm}
\label{tab:compare_with_r50_backbone}
}
\end{table*}

%% file: tables_tex/COCO_SwinL.tex
\begin{table*}[ht]
\centering\setlength{\tabcolsep}{7pt}
\renewcommand{\arraystretch}{1.2}
\footnotesize
\vspace{-3mm}
{
\resizebox{0.8\textwidth}{!}{%
\begin{tabular}{l|l|c|cccccc}
\shline
Model & Backbone & \#epochs  & AP & AP$_{50}$ & AP$_{75}$ & AP$_{S}$ & AP$_{M}$ & AP$_{L}$\\
\shline
$\mathcal{H}$-DETR~\cite{HDETR} & Swin-L (IN-$22$K) & $12$ & $56.1$ & $75.2$ & $61.3$ & $39.3$ & $60.4$ & $72.4$ \\
$\mathcal{H}$-DETR~\cite{HDETR} & Swin-L (IN-$22$K)  & $36$ & ${57.6}$ & ${76.5}$ & ${63.2}$ & ${41.4}$ & ${61.7}$ & ${73.9}$\\
Co-DETR~\cite{codetr2022} & Swin-L (IN-$22$K)  & $12$ & ${56.9}$ & ${75.5}$ & ${62.6}$ & ${40.1}$ & ${61.2}$ & ${73.3}$\\
DINO-4scale~\cite{zhang2022dino} & Swin-L (IN-$22$K) & $12$ & $56.8$ & $75.6$ & $62.0$ & $40.0$ & $60.5$ & $73.2$ \\
DINO-4scale~\cite{zhang2022dino} & Swin-L (IN-$22$K) & $36$ & $58.0$ & $\mathbf{77.1}$ & $\mathbf{66.3}$ & $41.3$ & $62.1$ & $73.6$ \\
\shline
\rowcolor{gray!15} Stable-DINO-4scale (ours) & Swin-L (IN-$22$K)  & $12$ & ${57.7}$ {(+$0.9$)} & $75.7$ & ${63.4}$ & $39.8$ & ${62.0}$ & ${74.7}$ \\
\rowcolor{gray!15} Stable-DINO-4scale (ours) & Swin-L (IN-$22$K)  & $24$ & $\mathbf{58.6}$ {(+$0.6$)*} & $76.7$ & ${64.1}$ & $\mathbf{41.8}$ & $\mathbf{63.0}$ & $\mathbf{74.7}$ \\
\shline
\end{tabular}}
\vspace{-0.2cm}
\caption{{Comparison to prior DETR variants on COCO \texttt{val2017} with Swin-Transformer Large backbones. * We compare our 24-epoch Stable-DINO with 36-epoch DINO here.}}
\label{tab:compare_with_swin_backbone}
}
\end{table*}

%% file: tables_tex/MaskDINO_COCO.tex
\begin{table*}[ht]
\centering\setlength{\tabcolsep}{7pt}
\renewcommand{\arraystretch}{1.2}
\footnotesize
\vspace{-3mm}
{
\resizebox{0.8\textwidth}{!}{%
\begin{tabular}{l|c|cc|ccccc}
\shline
Model  & \#queries  & Mask AP & Box AP & AP$_{50}^{mask}$ & AP$_{75}^{mask}$ & AP$_{S}^{mask}$ & AP$_{M}^{mask}$ & AP$_{L}^{mask}$\\
\shline
Mask2Former~\cite{BowenCheng2022Mask2FormerFV}  & $100$ & $38.7$ & $-$ & $59.8$ & $41.2$ & $18.2$ & $41.5$ & $59.8$ \\
MaskDINO~\cite{li2022mask}  & $300$ & $41.4$ & $45.7$ & $62.9$ & $44.6$ & $21.1$ & $44.2$ & $61.4$\\
\shline
\rowcolor{gray!15} Stable-MaskDINO (ours)  & $300$ & $\mathbf{42.1} (+0.7)$ & $\mathbf{47.0} (+1.3)$ & $\mathbf{63.4}$ & $\mathbf{45.7}$ & $\mathbf{21.9}$ & $\mathbf{44.5}$ & $\mathbf{62.2}$  \\
\shline
\end{tabular}}
\vspace{-0.2cm}
\caption{Results of Stable-MaskDINO compared with other state-of-the-art instance segmentation models on COCO \texttt{val2017}. All models trained with a ResNet-50 backbone for 12 epochs.}
\vspace{-0.1cm}
\label{tab:maskdino}}
\end{table*}

%% file: tables_tex/generalization.tex
\begin{table}[h]
\begin{center}
\renewcommand\arraystretch{1.2}
\resizebox{1.0\columnwidth}{!}{%
\begin{tabular}{
l@{\hskip9pt} |  
c |
c
c
c
}
\shline

 Model &  AP & AP$_{s}$ & AP$_{m}$ & AP$_{l}$  \\
\shline
 Deformable-DETR \cite{zhu2020deformable}  & 43.8 & 26.7 & 47.0 & 58.0 \\
 \rowcolor{gray!15} Stable-Deformable-DETR (Ours) & 45.1(+1.3) & 28.6 & 48.8 &  61.3 \\
\shline
 DAB-Deformable-DETR \cite{liu2022dabdetr}  & 44.2  & 27.5 & 47.1 & 58.6 \\
 \rowcolor{gray!15} Stable-DAB-Deformable-DETR (Ours)  & 45.2(+1.0) & 27.7 & 49.0 & 61.6 \\
\shline
 $\mathcal{H}$-DETR \cite{liu2022dabdetr} & 48.6 & 30.7 & 51.2 & 63.5 \\
 \rowcolor{gray!15} Stable-$\mathcal{H}$-DETR (Ours) & 49.2 (+0.6) & 32.7 & 52.8 & 64.9 \\
\shline
\end{tabular}}
\vspace{-0.3cm}
\caption{Effectiveness of our methods on other DETR variants. All models are trained with a ResNet-50 backbone for 12 epochs. The models with the prefix ``Stable'' use our proposed methods. }
\vspace{-0.4cm}
\label{table:generalization}
\end{center}
\end{table}

%% file: tables_tex/abltion.tex
\begin{table}[t]
\renewcommand{\arraystretch}{1.2}
\vspace{-3mm}
\begin{center}
\resizebox{1.0\columnwidth}{!}{%
\begin{tabular}{l|ccc|ccc}
\shline
Model Id & PSL & PMC & Memory Fusion & AP & AP$_{50}$ & AP$_{75}$\\
\shline
0 (baseline)&         &           & & 49.0 & 66.6 & 53.5 \\
1 (baseline + NMS) &         &           & & 49.2 & 66.8 & 54.0 \\
\hline
2 & \cmark  &    & & 49.8 & 66.7 & 54.5 \\
3 & \cmark  & \cmark    & & 50.2 & 66.7 & 55.0\\
4 & \cmark  & \cmark    & Simple Fusion & 50.2 & 66.7 & 55.0\\
5 & \cmark  & \cmark    & U-like Fusion & 50.3 & 66.6 & 55.0 \\
6 & \cmark  & \cmark    & Dense Fusion & \textbf{50.4} & \textbf{67.3} & \textbf{55.1} \\
7 &   &     & Dense Fusion & {49.4} & \textbf{67.3} & {54.1} \\
\shline
\end{tabular}}
\vspace{-0.3cm}
\caption{Ablations for different configurations. We use ``PSL'' and ``PMC'' for the position-supervised loss (Sec. \ref{sec:IOU_loss}) and the position-modulated cost in matching (Sec. \ref{sec:IOU_matching}).  For a fair comparison, we list the baseline DINO with NMS (Model Id 1). All models except model 0 are tested with a NMS.}
\vspace{-0.7cm}
\label{tab:ablation}
\end{center}
\end{table}

%% file: tables_tex/loss_design.tex
\begin{table}[h]
\renewcommand\arraystretch{1.2}
\begin{center}
\resizebox{0.85\columnwidth}{!}{%
\begin{tabular}{
l | l@{\hskip9pt} |  ccc
}
\shline
 Model Id & $f_1(s, p)$ & AP & AP$_{50}$ & AP$_{75}$  \\
\shline
   0 (Model 1 in Table \ref{tab:ablation}) & $1$  & 49.2 & 66.8 & 54.0  \\ 
\hline
   1 & $s^{0.5}$  & 49.3 & \textbf{67.5} & 53.7 \\
   2 & $s$  & 49.4 & 67.2 & 54.2 \\
   3 & $s^2$  & \textbf{49.6} & 66.8 & \textbf{54.5} \\
   4 & $s^3$  & 49.4 & 65.8 &  54.3 \\
\hline
   5 &  $s^1p^{0.25}$  & 48.6 & 66.0 & 52.8  \\    
   6 &  $s^1p^1$  & 26.4 & 32.5 & 28.8 \\    
   7 &  $s^2p^1$  & 27.4 & 33.7 & 29.8 \\    
\hline
   8 &  $(e^{s}-1)/(e-1)$  & \textbf{49.6} & 66.8 & 54.4 \\        
   9 &  $\sin(s \times \pi / 2)$  & 48.5 & 67.3 & 52.8 \\       
\shline
\end{tabular}}
\vspace{-0.3cm}
\caption{Ablations for different loss designs for the position-supervised loss. $s$ and $p$ are used for IOU scores and classification probabilities, respectively. $f_1(s,p)$ is an extended function of the $f_1(s)$ in Eq. \ref{eq:new_loss} to include classification probabilities. }
\vspace{-0.6cm}
\label{table:loss_design}
\end{center}
\end{table}

%% file: tables_tex/loss_weight.tex
\begin{table}[h]
\begin{center}
\renewcommand\arraystretch{1.2}
\resizebox{0.85\columnwidth}{!}{%
\begin{tabular}{
l | l@{\hskip9pt} |  ccc
}
\shline
Model Id & CLS Weight & AP & AP$_{50}$ & AP$_{75}$  \\
\shline
    0 & 4.0 & 49.7 & 66.3 & 54.5\\
    1 & 5.0 & 50.1 & 66.6 & 54.9\\    
    2 (Model 3 in Table \ref{tab:ablation}) & 6.0 & \textbf{50.2} & \textbf{66.7} & \textbf{55.0}\\   
    3 & 8.0 & 49.9 & 66.5 & 54.7\\   
    4 & 10.0 & 49.6 & 66.5 & 54.5\\       
    5 & 20.0 & 48.3 & 65.9 & 52.7\\   
\shline
\end{tabular}}
\vspace{-0.3cm}
\caption{Ablations for different loss weights for the position-supervised loss. The ``CLS weight'' means the classification weight in final losses.}
\vspace{-0.5cm}
\label{table:loss_weight}
\end{center}
\end{table}

%% file: tables_tex/ablation_cost.tex
\begin{table}[h]
\renewcommand\arraystretch{1.2}
\begin{center}
\resizebox{0.9\columnwidth}{!}{%
\begin{tabular}{
l | l@{\hskip9pt} | l@{\hskip9pt} | ccc
}
\shline
 Model Id & $f_2(s)$ & Cost Weight & AP & AP$_{50}$ & AP$_{75}$  \\
\shline
   0 (Model 2 in Table \ref{tab:ablation}) & $1$ & 2.0 & 49.8 & 66.7 & 54.5  \\ 
\hline
   1 & $s^{0.25}$  & 2.0 & 50.0 & 66.7 & 54.9 \\
   2 & $s^{0.5}$  & 2.0 & \textbf{50.2} & {66.7} & \textbf{55.0} \\
   3 & $s$  & 2.0 & 49.7 & 65.8 & 54.7 \\
   4 & $s^2$  & 2.0 & 48.8 & 64.7 & 53.7 \\
\hline
   5 & $s^{0.5}$  & 1.0 & 49.6 & 64.6 & 55.0 \\
   6 & $s^{0.5}$  & 4.0 & 49.6 & \textbf{67.4} & 54.0 \\ 
   7 & $s^{0.5}$  & 8.0 & 48.8 & 67.1 & 52.7 \\ 
\shline
\end{tabular}}
\vspace{-0.3cm}
\caption{Ablations for different designs and cost weight for the position-modulated cost. $s$ means the IOU scores. $f_2(s)$ is the function defined in Eq. \ref{eq:cls_cost}. }
\vspace{-0.6cm}
\label{table:abla_cost}
\end{center}
\end{table}

%% file: images_tex/compare_top30.tex
\begin{figure}[h]
\begin{center}
   \includegraphics[width=\linewidth]{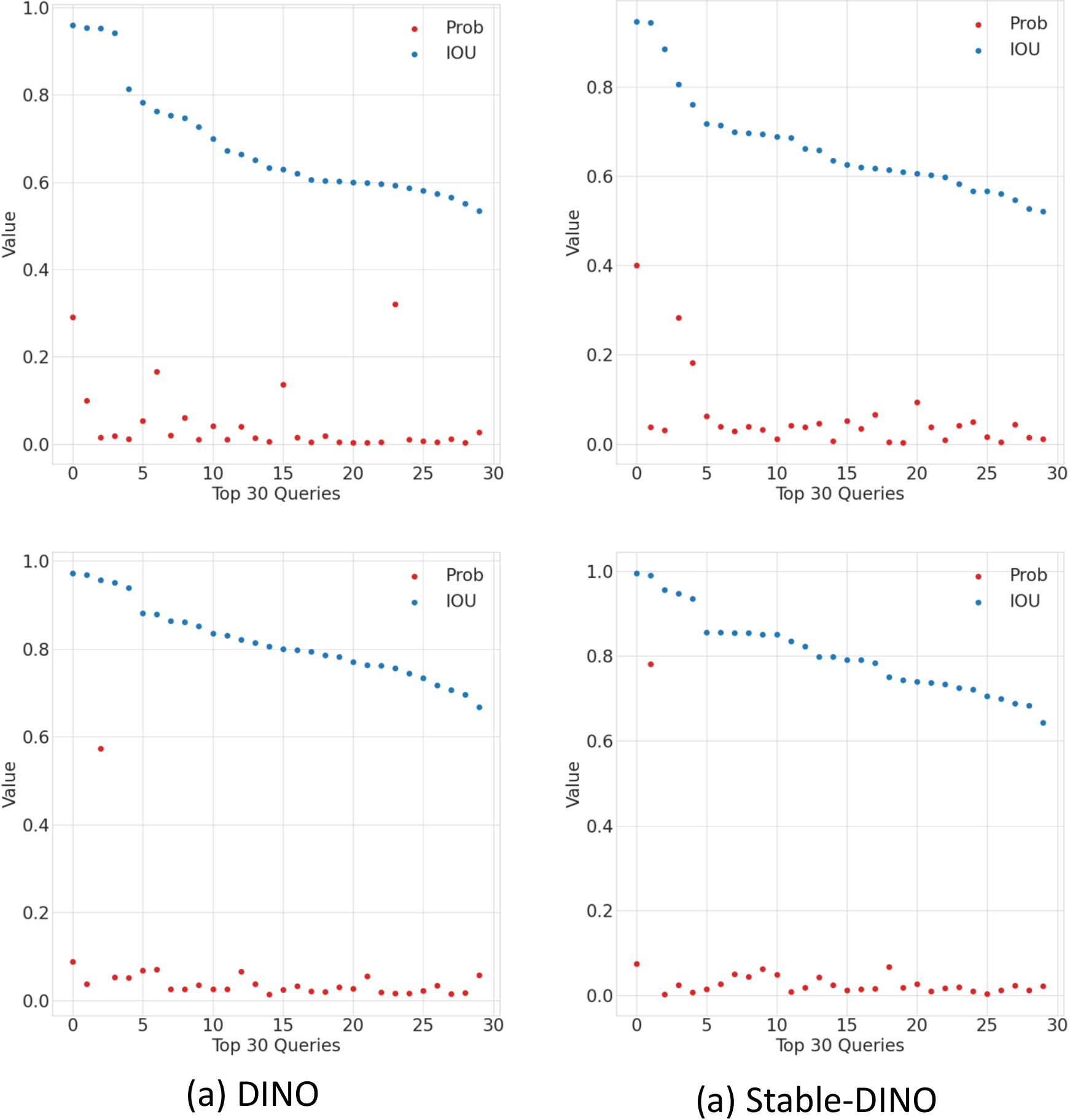}
\end{center}
\vspace{-0.3cm}
   \caption{Comparison of top 30 queries with highest IOU values in DINO (a) and Stable-DINO (b). }
\label{fig:compare_top30}
\end{figure}

%% file: images_tex/memory_fusion.tex
\begin{figure}[t]
    \centering
    \includegraphics[width=0.8\linewidth]{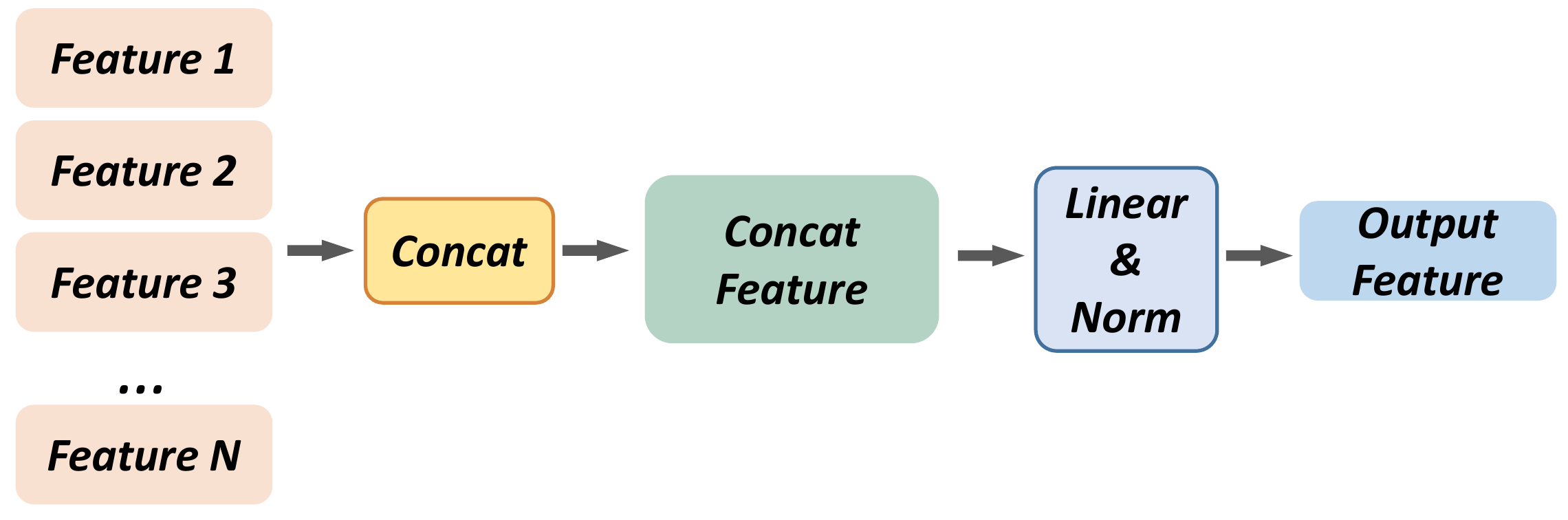}
    \vspace{-0.1cm}
    \caption{Detailed operation of memory fusion.}
    \label{fig:memory_fusion}
\end{figure}

%% file: images_tex/compare_dino_stabledino_convergence.tex
\begin{figure}[h]
\begin{center}
   \includegraphics[width=\linewidth]{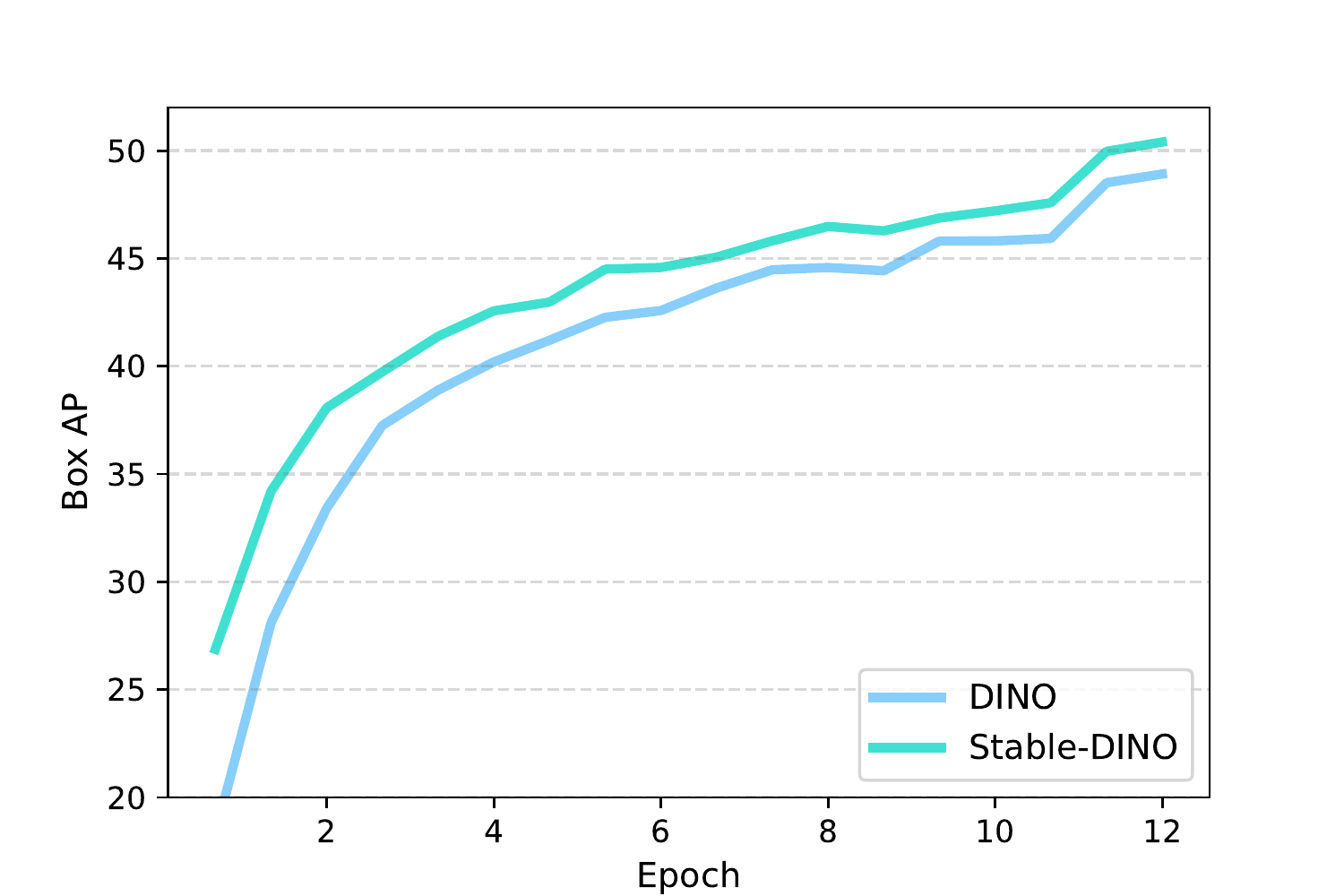}
\end{center}
   \caption{Convergence comparison of DINO and Stable-DINO.}
\label{fig:convergence_dino_stabledino}
\end{figure}

%% file: images_tex/comapre_sampling_points.tex
\begin{figure*}[t]
\begin{center}
   \includegraphics[width=\linewidth]{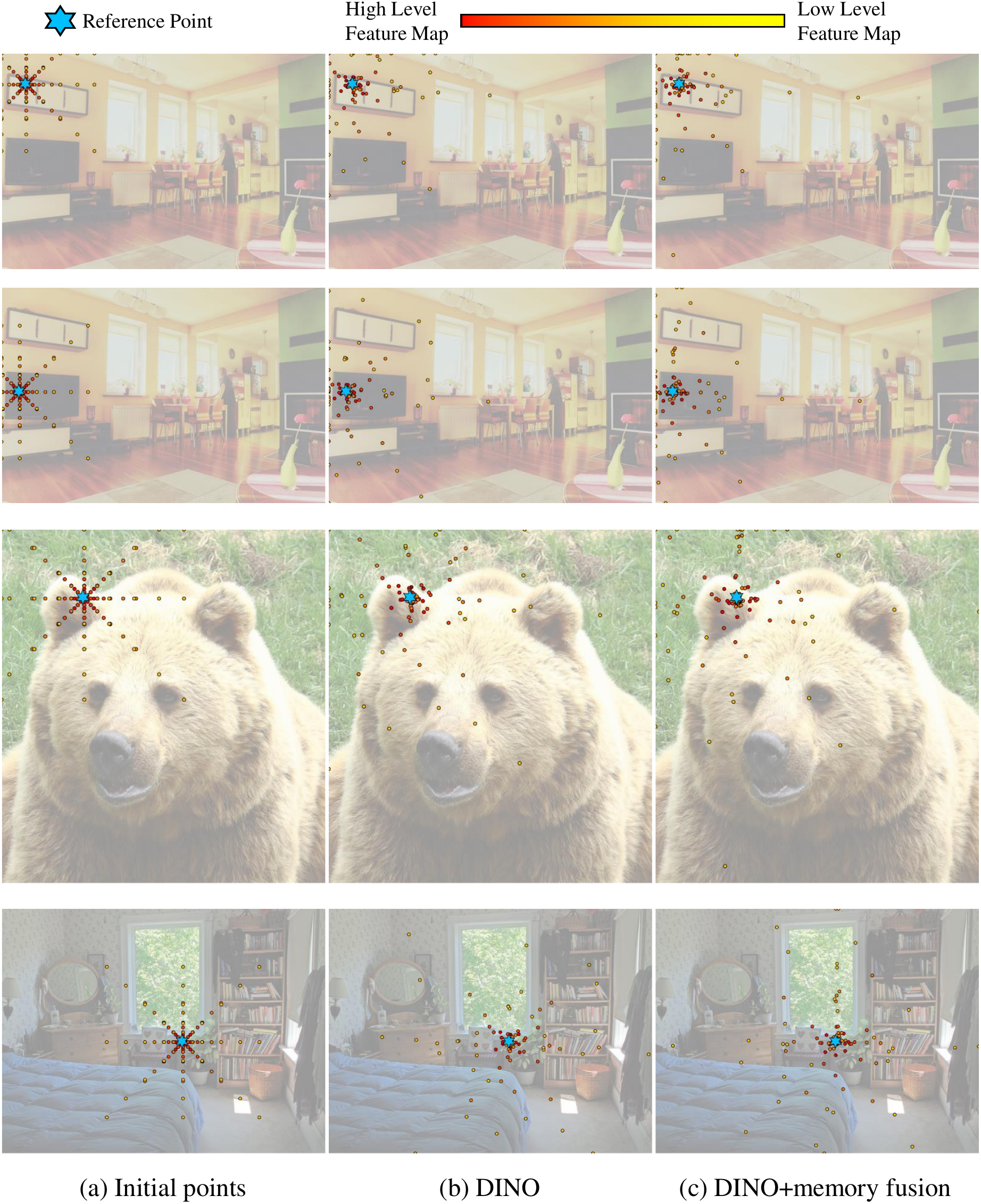}
\end{center}
\vspace{-0.3cm}
   \caption{Comparisons of sampling points.}
\label{fig:compare_sampling_points}
\end{figure*}

%% file: images_tex/vis_1.tex
\begin{figure*}[t]
\begin{center}
   \includegraphics[width=\linewidth]{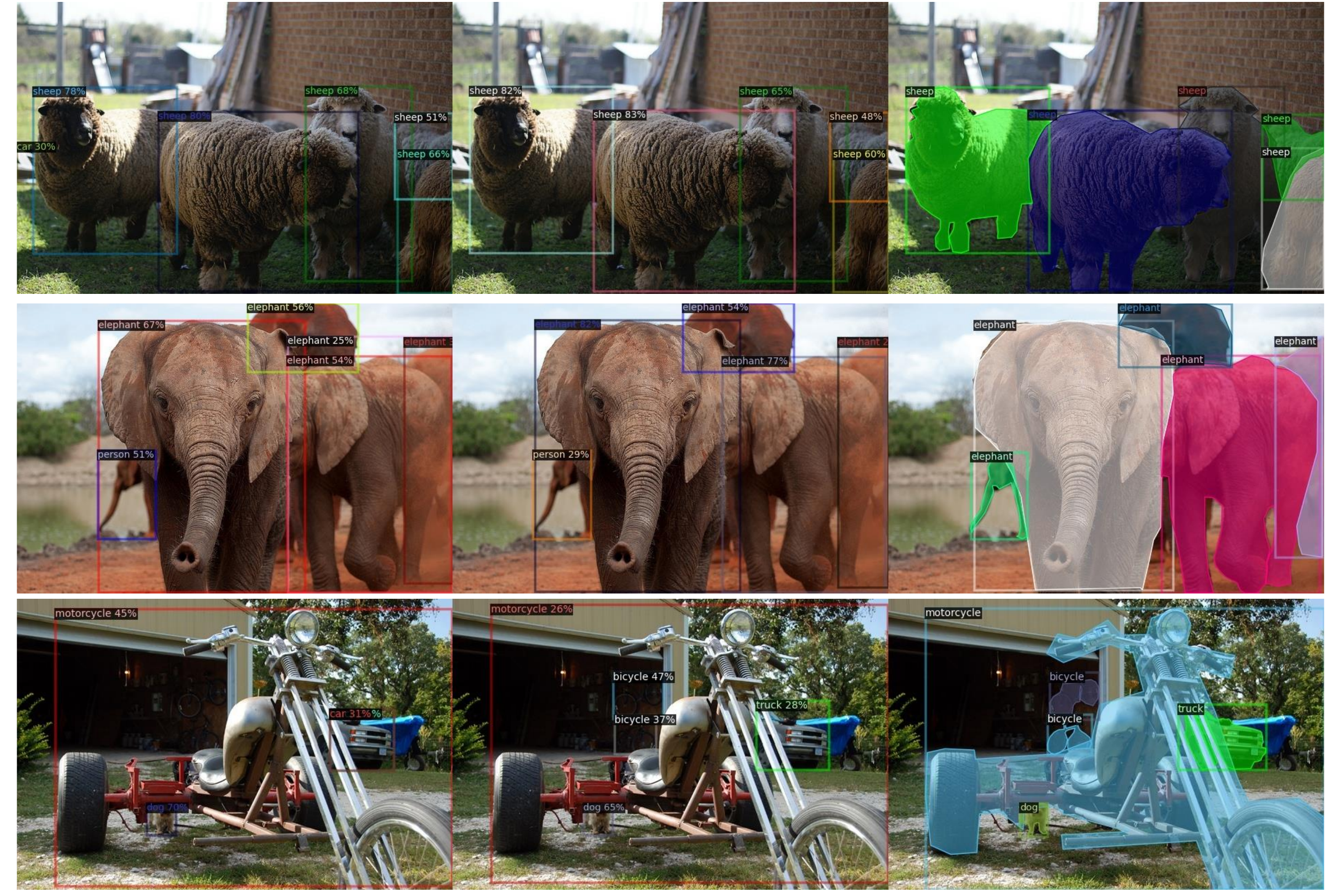}
   \includegraphics[width=\linewidth]{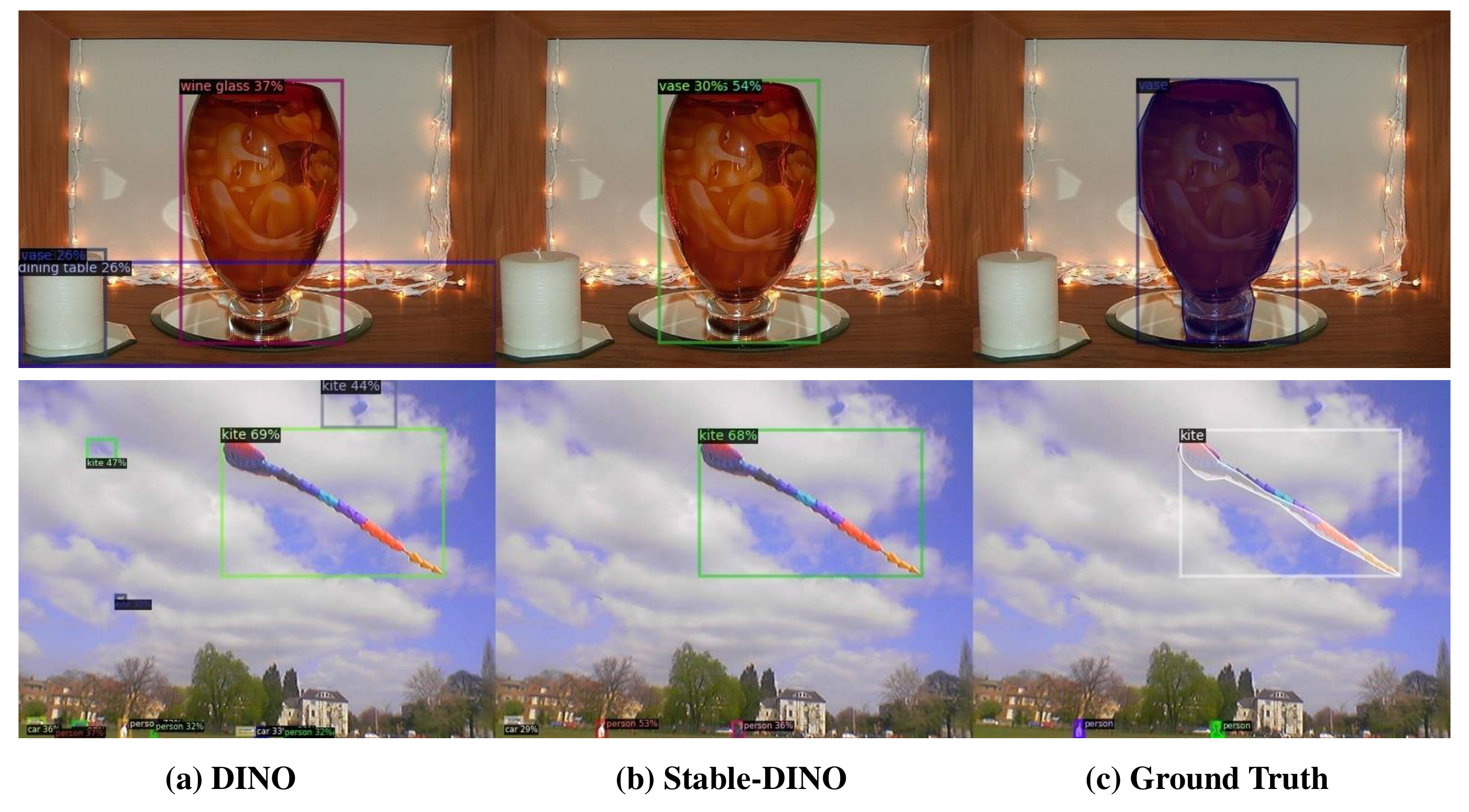}
\end{center}
\vspace{-0.3cm}
   \caption{We compare the results of DINO (a) and Stable-DINO (b). The column (c) presents the ground truth results.}
\label{fig:vis_1}
\end{figure*}